\documentclass[journal]{IEEEtran}
\ifCLASSINFOpdf
\else
\fi
\usepackage{times}
\usepackage{epsfig}
\usepackage{graphicx}
\usepackage{array}
\usepackage{algorithm}
\usepackage{algorithmic}
\usepackage{microtype}
\usepackage{graphicx}
\usepackage{subfigure}
\usepackage{booktabs} 
\usepackage{multirow}
\usepackage{array}
\usepackage{amsmath}
\usepackage{amssymb}

\newcolumntype{I}{!{\vrule width 3pt}}
\newlength\savewidth
\newcommand\shline{\noalign{\global\savewidth\arrayrulewidth
                            \global\arrayrulewidth 1.5pt}%
                   \hline
                   \noalign{\global\arrayrulewidth\savewidth}}

\usepackage[pagebackref=true,breaklinks=true,letterpaper=true,colorlinks,bookmarks=false]{hyperref}

\begin{document}
%
\title{Model Inconsistent but Correlated Noise:\\
Multi-view Subspace Learning with Regularized Mixture of Gaussians}
%
%
%

\author{Hongwei~Yong,
        Deyu~Meng,
        Jinxing~Li,
        Wangmeng~Zuo,
        Lei~Zhang
\thanks{H. Yong, J. Li and L. Zhang are with the Department of Computing, the
Hong Kong Polytechnic University, Hong Kong. (e-mail: cshyong@comp.polyu.edu.hk; csjxli@comp.polyu.edu.hk; cslzhang@comp.polyu.edu.hk)}
%
\thanks{D. Meng (corresponding author) is with the School of Mathematics and Statistics and Ministry
of Education Key Lab of Intelligent Networks and Network Security, Xi'an
Jiaotong University. (e-mail: dymeng@mail.xjtu.edu.cn)}
\thanks{W. Zuo is with the School of Computer Science and Technology,
Harbin Institute of Technology, Harbin 150001, China. (e-mail:
cswmzuo@gmail.com)}

}

\maketitle


\begin{abstract}
Multi-view subspace learning (MSL) aims to find a low-dimensional subspace of the data obtained from multiple views. Different from single view case, MSL should take both common and specific knowledge among different views into consideration. To enhance the robustness of model, the complexity, non-consistency and similarity of noise in multi-view data should be fully taken into consideration. Most current MSL methods only assume a simple Gaussian or Laplacian distribution for the noise while neglect the complex noise configurations in each view and noise correlations among different views of practical data. To this issue, this work initiates a MSL method by encoding the multi-view-shared and single-view-specific noise knowledge in data. Specifically, we model data noise in each view as a separated Mixture of Gaussians (MoG), which can fit a wider range of complex noise types than conventional Gaussian/Laplacian. Furthermore, we link all single-view-noise as a whole by regularizing them by a common MoG component, encoding the shared noise knowledge among them. Such regularization component can be formulated as a concise KL-divergence regularization term under a MAP framework, leading to good interpretation of our model and simple EM-based solving strategy to the problem. Experimental results substantiate the superiority of our method.
\end{abstract}

\section{Introduction}
\begin{figure}[!]
\centering
\centering
\includegraphics[width=0.48\textwidth]{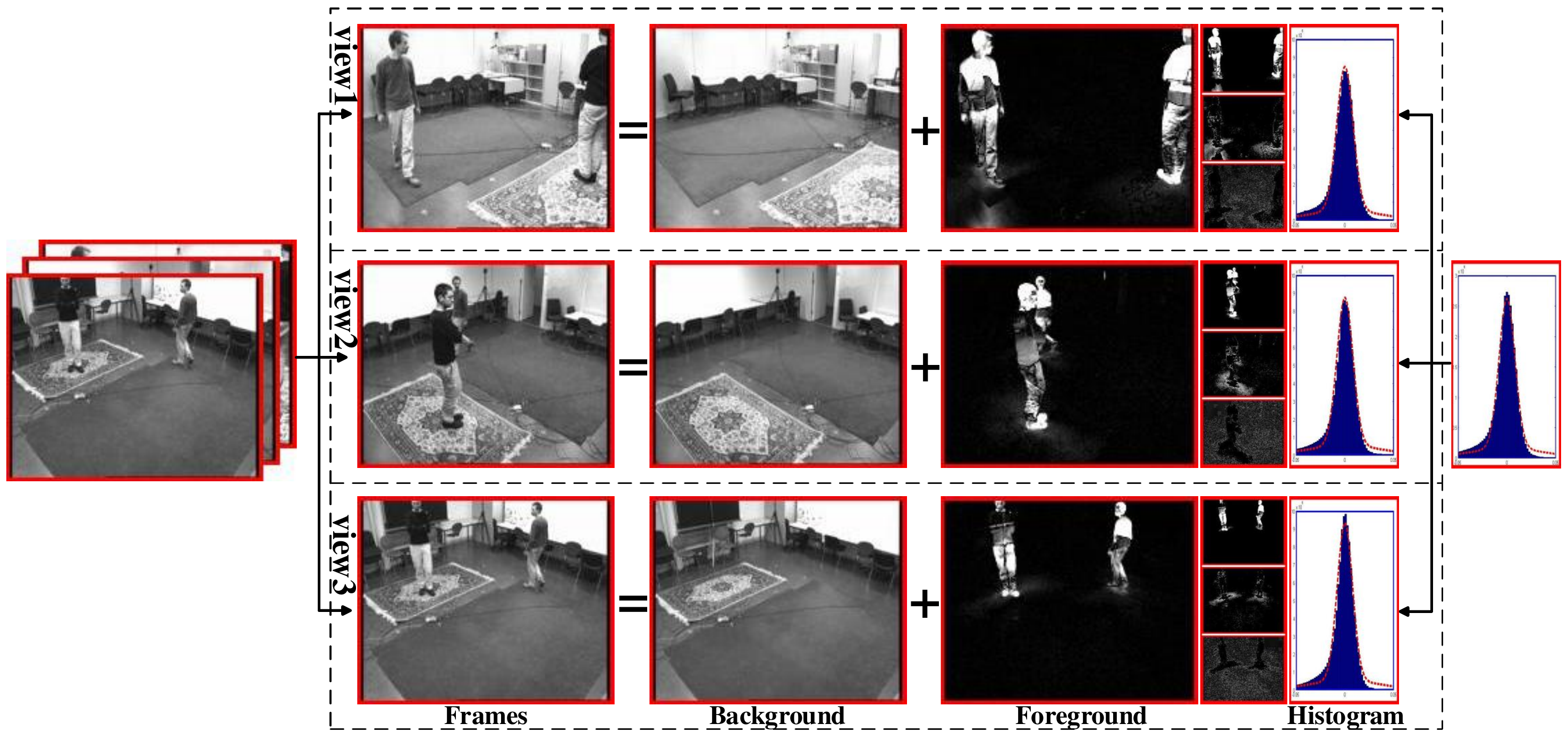}
\label{fig2:video}\vspace{-0mm}
   \caption{ MSL-RMoG performance on videos collected through multiple survivance cameras. From left to right: original frame from each view, the low-rank background, the residual of each view, three Gaussian components alongside each residual extracted by MSL-RMoG, and the histograms of residuals of each view. The most righthand panel shows the histogram of the common residual shared among all three residuals. It is easy to see that each view has its specific noise configuration, with evident correlation between each other. }
\label{fig2}\vspace{-0mm}
\end{figure}

An increasing number of applications, including face recognition, video surveillance, social computing and 3-D point cloud reconstruction, require the data obtained from various domains or extracted from diverse feature extractors to achieve a high accuracy and satisfactory performance. These kinds of data are known as multi-view data. For instance, videos can be generated from different angles (as shown in Fig. \ref{fig2}) or from different sensors in a surveillance scene, and a given image can be represented by different types of features such as SIFT and HoG.

It has been proven that these data are more comprehensive and sufficient to indicate a particular object or situation than that obtained from only a single view \cite{xu2015multi}, as a result of more abundant information. Thus the multi-view related research plays an important role in both academic and practical fields. Various multi-view learning methods have been presented for both supervised and unsupervised learning problems. For the former, multi-view based face recognition \cite{xu2015multi} and video tracking \cite{Berclaz11}, \cite{Fleuret08a} have achieved good performance in practical applications. For the latter, multi-view data restoration and recovery, or multi-view subspace learning (MSL) have been proposed in \cite{white2012convex}, \cite{jia2010factorized} etc. This paper mainly focuses on the unsupervised MSL issue, which is an important branch in this research line.
The basic assumption of MSL is that each specific view of data lies on a low-dimensional subspace, and these multi-view subspaces share some common knowledge. Through properly encoding and learning such multi-view common knowledge, the subspace learning task for each view can be finely compensated and extracted more appropriately than that only using one simple view knowledge. Most current MSL methods encode such common knowledge by the deterministic components of data, like the shared subspace \cite{ding2014low,zhushared}, or similar coefficient representations of different views \cite{guo2013convex,lock2013joint,xu2015multi,white2012convex,xu2017cross,ma2016decorrelation}. Such assumption is always rational in problems and can get satisfactory performance in cases where the contexts and scenarios of different views are consistent and not very complicated.

There are, however, still critical limitations of current methods, especially when being used on multi-view data with complicated noisy scenarios. Firstly, the current methods generally use a simple $L_2$-norm or $L_1$-norm loss in the model, implying that they assume the noise embedded in each view of data following a simple Gaussian or Laplacian. In practical cases, however, the noise is always much more complicated, like those shown in Fig. \ref{fig2}. E.g., there may possibly exist foreground objects, along with their shadows and weak camera noises, in a multi-view surveillance video (see Fig. \ref{fig2:video}). Such complicated noise evidently cannot be finely approximated by a simple Gaussian/Laplacian noise as traditional, which always degenerates their robustness in practical applications.

Secondly, most current methods utilize an unique loss-term to encode all views of data, which implicitly assumes an i.i.d. noise among all data. This, however, is always not correct in many real cases. The noises in different views of data are always of evident distinctiveness due to their different collecting angles, domains and sensors. E.g., in multi-view videos collected from different surveillance cameras, some views might capture a foreground object with big occlusion area, which makes the noise in the view should be better encoded as a long-tailed distribution like Laplacian (i.e., better using $L_1$-norm loss), while other views might just miss such object, which makes the view contains weak noise signals and leads to a better Gaussian approximation (i.e., better using $L_2$-norm loss), as clearly shown in Fig. \ref{fig2}. The neglecting of such noise distinctiveness among different views tends to negatively influence the performance of current methods.

Last but not least, besides distinctiveness, there is also correlation and  similarity among noises in different views of data. E.g., when one view of videos has large occlusion noise, implying there is an evident object entering the surveillance area, thus, more other views might possibly have large noises, and should be commonly encoded with long-tailed distributions. We should consider such noise similarity among all views to further enhance the noise fitting capability as well as robustness of the multi-view learning strategy.

To address the aforementioned noise fitting issue, in this work we initiate a MSL method by fully
taking the complexity, non-consistency and similarity of noise in multi-view learning
into consideration . To our best knowledge, this is the first work to consider stochastic components in multi-view learning in such an elaborate manner to make it robust to practical complicated noises. The main contributions can be summarized as follows:
\begin{itemize}
\item
To address the problem of modeling such intra-view complicated and  inconsistent while inter-view correlated noise, we apply this KL divergence regularization into the noise modeling of multi-view subspace learning by formulating each view a separated MoG for its noise and regularizing them with KL divergence term instead of only using an uniform MoG as conventional.

\item Further, we propose an EM algorithm to solve the model with KL divergence regularization, and each involved step can be solved efficiently. To be more specific, all of the parameters and variables of noise distribution have a closed-form solution in M step.
\item
A detailed theoretical explanation is given for KL divergence regularization
by conjugate prior for local distribution and KL  divergence average
for global  distribution. Moreover, to utilize this regularization term into complex noise modeling succinctly, we extend it to a joint form for mixture of fully exponential family distributions (including MoG) by using a certain alternative regularization term which is a upper bound of original term.

%
%
%
    \end{itemize}

The paper is organized as follows: Section 2 reviews some related works about MSL. Section 3 proposes our model and Section 4 designs an EM algorithm for solving the model. Section 5 presents some theoretical explanations on KL divergence regularization used in the model. Section 6 gives experiments and finally a conclusion is made.

\section{Related works}
\vspace{-5pt}
In recent years, numbers of multi-view learning approaches have been proposed. The Canonical Correlation Analysis (CCA) \cite{hotelling1936relations} which learns a shared latent subspace across two views is a typical method to analyze the linear correlation. To handle nonlinear alignment, the kernel CCA \cite{akaho2006kernel} was proposed by projecting data into a high-dimensional feature space. Additionally, the sparsity as a prior distribution was imposed to CCA \cite{archambeau2009sparse}. Several robust CCA-based strategies were proposed by Nicolaou et al.  \cite{nicolaou2014robust} and Bach et al.  \cite{bach2005probabilistic}. The $L_{1}$-loss was introduced to limit the influence of outliers and noise in \cite{nicolaou2014robust}, while a Student-$t$ density model was presented in \cite{bach2005probabilistic} to handle outliers.

Other works on MSL have also attracted much attention recently. \cite{jia2010factorized} used structured sparsity to deal with multi-view learning problem by solving two convex optimization alternately. Similarly, Guo~\cite{guo2013convex} proposed a Convex Subspace Representation Learning (CSRL) for multi-view subspace learning; this technique relaxed the problem and reduced dimensionality while retaining a tractable formulation. Other related methodologies on convex formulations for MSL can be found in \cite{guo2013convex,goldberg2010transduction,cabral2011matrix,christoudias2012multi,behmardi2014overlapping} . Moreover, a Gaussian process regression \cite{shon2005learning} was developed to learn a common nonlinear mapping between corresponding sets of heterogenous observations. Multiple kernel learning (MKL) \cite{gonen2011multiple,xu2010simple} has also been widely used for multi-view data since combining kernels either linearly or nonlinearly has a crucial improvement on learning performance. The works in \cite{xu2015multi} and \cite{white2012convex}
exploited the Cauchy loss \cite{mizera2002breakdown} and $L_1$ loss, respectively, to strengthen robustness to the noise. Considering correlation and independence of multi-view data, several methods have been introduced to divide the data into correlated components to all views and specific components to each view. Lock et al.  \cite{lock2013joint} presented a Joint and Individual Variation Explained (JIVE) which decomposes the multi-view data into three parts: a low-rank subspace capturing common components, a low-rank subspace obtaining individual ones and a residual errors. Inspired by JIVE, Zhou et al. \cite{zhou2015group} used a common orthogonal basis extraction (COBE) algorithm to identify and separate the shared and individual features.

However, the conventional methods put their main focus on the deterministic components of multi-view data, while not elaborately consider the complicated stochastic components (i.e., noises) in data, which inclines to degenerate their robustness especially in real cases with complex noise.
To the best of our knowledge, our method is the first one investigating this MSL noise issue and formulating the model capable of both adapting intra-view noise complexity (by parametric MoG) and delivering inter-view noise correlation (by KL-divergence regularization). Its novelty reflects in both its investigated issue and designed methodology (regularized noise modeling).

\section{Problem Formulation}

\subsection{Notation}
The observed multi-view data is denoted as $\mathbf{X}=\{\mathbf{X}^{v}\}_{v=1}^V$, where $\mathbf{X}^{v}\in \mathbb{R}^{m\times n} $ means the data collected from the $v^{\text{th}}$ view, $m,n$ mean the dimension and number of data in each view\footnote{To notion convenience, we assume each view has the same data dimensionality and number. Yet our method can also be easily used in cases where they are different in multiple views.}. We set
$\boldsymbol{\theta}^{model}=
\{ \{ \mathbf{L} ^{v} \}_{v=1}^V,\{ \mathbf{R} ^{v} \}_{v=1}^V, \mathbf{R}  \}$ as the variables in deterministic parts of our model, where $\mathbf{L} ^{v} \in \mathbb{R}^{m\times r} $, $ \mathbf{R} ^{v} ,\mathbf{R}\in \mathbb{R}^{n\times r}$ ($r\ll\text{min}(m,n)$ is the subspace rank) denote the subspace parameters. Besides, denote
$\boldsymbol{\theta}^{MoG}=\{\left \{ \mathbf{\Pi} ^{v} \right \}_{v=1}^V$, $\left \{ \mathbf{\Sigma} ^{v} \right \}_{v=1}^V,\mathbf{\Pi} $, $ \mathbf{\Sigma}\}$ as variables in stochastic parts of our model, where $\mathbf{\Pi} ^{v}=\{ \pi _{k}^v \}_{k=1}^K$, $\mathbf{\Sigma} ^{v}=\{ \sigma _{k}^v \}_{k=1}^K$, $\mathbf{\Pi} =\{ \pi _{k} \}_{k=1}^K$ as well as $\mathbf{\Sigma} =\{ \sigma _{k} \}_{k=1}^K$. Denote $\boldsymbol{\theta}=\{\boldsymbol{\theta}^{model},\boldsymbol{\theta}^{MoG}\}$ as all variables involved in our model. The parameters
 $\lambda$ , $\lambda_L$, and $\lambda_R$ denote the hyperparameters to infer model variables. $\mathbf{A}_{i \cdot}$ and $\mathbf{A}_{\cdot i}$ means the $i^{th}$ row and column vectors of matrix $\mathbf{A}$, respectively. $N_{\pi}$ and $N_{\sigma_k}$ are the strength parameters of regularization term.

\subsection{Probabilistic Modeling}

\subsubsection{MoG modeling on each view of noises}
As conventional MSL method \cite{lock2013joint}, we model deterministic component of each view of data as a specific subspace $\mathbf{L}^{v}$ with specific coefficients $\mathbf{R}^{v}$ supplemented a shared coefficients $\mathbf{R}$ among all views. Then each element $x_{ij}^{v}$ ($i=1,2\cdots ,m$, $j=1,2\cdots ,n$) of the input matrix $\mathbf{X}^{v}$ is modeled as:
\vspace{-0pt}
\begin{equation}\small
\begin{aligned}
x_{ij}^{v}=(\mathbf{L}_{i \cdot}^{v})^{T}(\mathbf{R}_{\cdot j}+\mathbf{R}_{\cdot j}^{v})+e_{ij}^{v},
\end{aligned}\label{comp}
\end{equation}
where $e_{ij}^{v}$ represents the noise in $x_{ij}^{v}$. From Eq. (\ref{comp}) we can see that a shared variable $\mathbf{R}$ is learned to reveal the relationship among different views.

Unlike previous works \cite{nicolaou2014robust}, \cite{panagakis2015robust}, \cite{huber2011robust} only using a simple Gaussian or Laplacian to model noise distribution, we model $e_{ij}^{v}$ in each view as a MoG distribution to make it better adapt the noise complexity in practice \cite{maz1996approximate}. I.e.,
\vspace{-0pt}
\begin{equation}\small
p(e_{ij}^{v}\mid \mathbf{z}_{ij}^{v} ,\mathbf{\Sigma}^{v} )=\prod_{k=1}^{K}\mathcal{N}(e_{ij}^{v}\mid 0,{\sigma _{k}^v}^{2})^{z_{ijk}^v},
\end{equation}\vspace{-8pt}

\noindent where $p(\mathbf{z}_{ij}^v|\mathbf{\Pi}^{v})=\text{Multi}(\mathbf{z}_{ij}^v|\mathbf{\Pi}^{v})$, $\mathbf{z}_{ij}^v$ denotes the latent variable , which satisfies $z_{ijk}^v\in\{0,1\}$ and $\sum_{k=1}^K {z}_{ijk}^v =1 $, ``Multi" denotes the multinomial distribution. Note that the MoG parameters $\mathbf{z}_{ij}^{v} ,\mathbf{\Sigma}^{v}$ are different from view to view, implying that each view has its specific noise configuration.

\begin{figure}[t]
\begin{center}
 \includegraphics[width=0.9\linewidth]{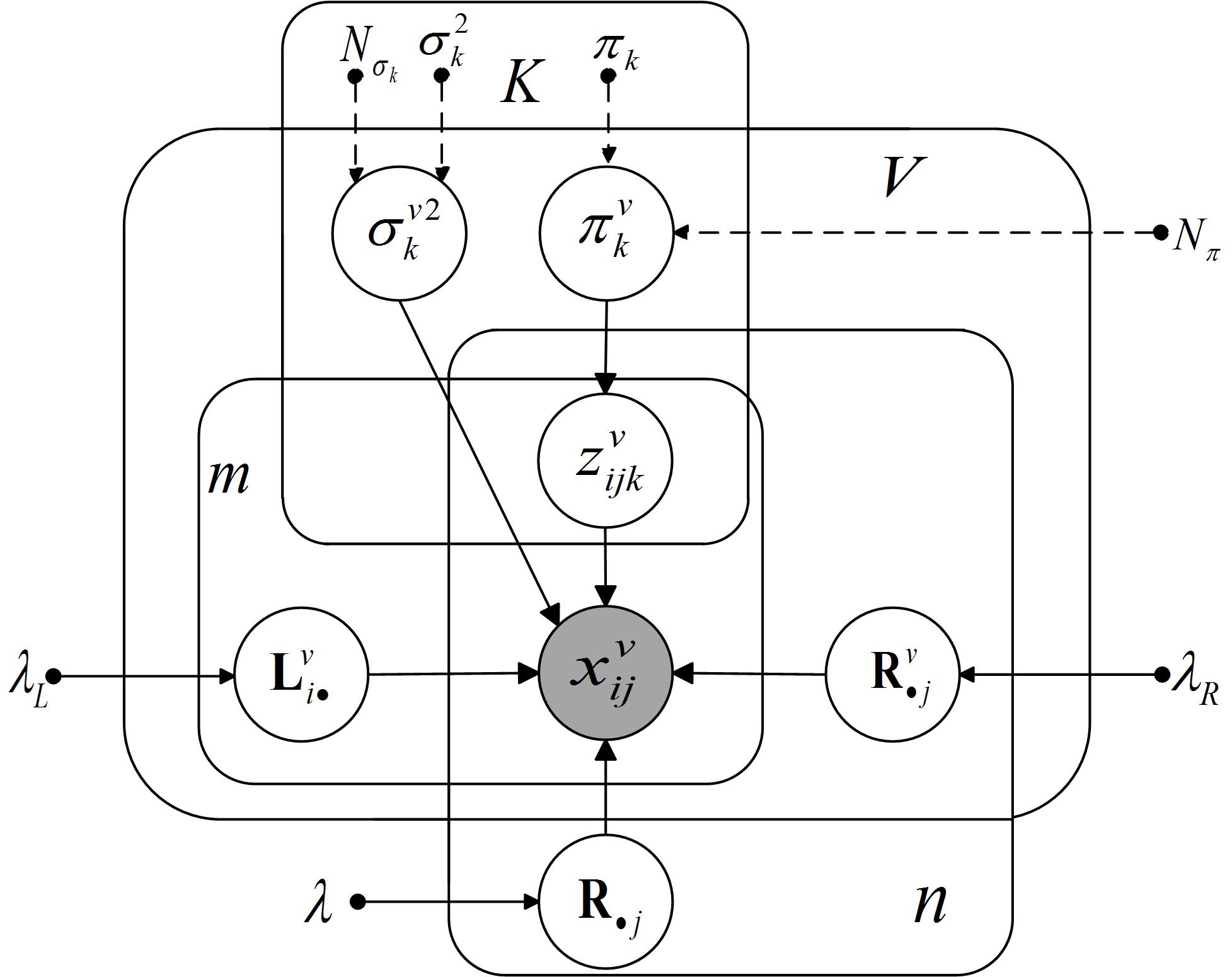}
\end{center}\vspace{-0mm}
   \caption{The graphical model of MSL-RMoG}
\label{fig:GraphicalModel}
\end{figure}

Then we can deduce that:
\vspace{-0pt}
\begin{equation}\small
\begin{aligned}
&p(x_{ij}^{v}\mid \mathbf{L}_{i \cdot}^{v},\mathbf{R}_{\cdot j},\mathbf{R}_{\cdot j}^{v},\mathbf{z}_{ij}^{v},\mathbf{\Sigma}^{v} )=\\
&\ \ \ \ \ \prod_{k=1}^{K}\mathcal{N}(x_{ij}^{v}\mid(\mathbf{L}_{i \cdot}^{v})^{T}(\mathbf{R}_{\cdot j}+\mathbf{R}_{\cdot j}^{v}),{\sigma _{k}^v}^{2})^{z_{ijk}^v}\\
&p(\mathbf{z}_{ij}^v|\mathbf{\Pi}^{v})=\text{Multi}(\mathbf{z}_{ij}^v|\mathbf{\Pi}^{v}); p(\mathbf{L}_{i \cdot}^{v}\mid \lambda _{L})\sim \mathcal{N}(0,\frac{1}{\lambda _{L}}\mathbf{I});\\
&p(\mathbf{R}_{\cdot j}^{v}\mid \lambda _{R})\sim \mathcal{N}(0,\frac{1}{\lambda _{R}}\mathbf{I}); p(\mathbf{R}_{\cdot j}\mid \lambda )\sim \mathcal{N}(0,\frac{1}{\lambda}\mathbf{I}).
\end{aligned}
\end{equation}
The prior distributions on $\mathbf{L}^v$, $\mathbf{R}$ and $\mathbf{R}^{v}$ are to constrain their scales to avoid overfitting problem. The full graphical model of our proposed method is shown in Fig. \ref{fig:GraphicalModel}.

\subsubsection{Shared MoG modeling on all views of noises}

In order to encode the correlation of noises among different views of data, we assume that there is
 prior distribution $p(\boldsymbol{\theta}^{MoG})$ to $\boldsymbol{\theta}^{MoG}$, which is related to a KL divergence regularization term. Then all of the model variables and MoG parameters can be inferred by the MAP estimation. After marginalizing the variable $\mathbf{z}_{ij}^v$, the posterior of $\mathbf{X}$ then can be written as:
 \vspace{-0pt}
\begin{equation}
\begin{aligned}
p(\boldsymbol{\theta})\varpropto
 p(\mathbf{X}|\boldsymbol{\theta})
 p(\boldsymbol{\theta}^{model}|\lambda, \lambda_L, \lambda_R)p(\boldsymbol{\theta}^{MoG}).
\end{aligned}\label{MAP}
\end{equation}

Then our method needs to solve the problem of minimizing the following objective (negative log of Eq. (\ref{MAP})):
\vspace{-0pt}
\begin{equation}\small
\begin{aligned}
\mathcal{L}(\boldsymbol{\theta})=
-\ln p(\mathbf{X}|\boldsymbol{\theta})+
\mathcal{R}(\boldsymbol{\theta}^{model})+\mathcal{R}(\boldsymbol{\theta}^{MoG}),
\end{aligned}\label{obj_fun}
\end{equation}
where
\vspace{-0pt}
\begin{equation}\small
\begin{aligned}
&\ln p(\mathbf{X}|\boldsymbol{\theta})=
\sum_{v=1}^{V} \sum_{i,j\in \Omega^{v}} \ln \sum_{k} \pi _{k}^{v}\mathcal{N}(x_{ij}^{v}\mid \mathbf{ L}_{i \cdot}^{v}(\mathbf{R}_{\cdot j}^{v}+\mathbf{R}_{\cdot j}),{\sigma_k^v}^2)\\
&\mathcal{R}(\boldsymbol{\theta}^{model})=\lambda \left \| \mathbf{R} \right \|_{F}^{2}
+\sum_{v=1}^{V}(\lambda _{R}\left \| \mathbf{R}^{v} \right \|_{F}^{2}+\lambda _{L}\left \| \mathbf{L}^{v} \right \|_{F}^{2}),
\end{aligned}\label{obj_each}
\end{equation}
where $\Omega^{v}$ is the index set of the non-missing entries in $\mathbf{X}^{v}$.

For regularization term $\mathcal{R}(\boldsymbol{\theta}^{MoG})$, we can easily use the following KL divergence form:
\vspace{-0pt}
\begin{equation}\small
\begin{aligned}&\mathcal{R}(\boldsymbol{\theta}^{MoG})=\sum_{v=1}^{V}\{ N_{\pi}D_{KL}(\text{Multi}(\boldsymbol{\Pi})||\text{Multi}(\boldsymbol{\Pi}^v))\\ &\ \ \ \ \ \ \ \ \ \ \ \ \ \ \ \ \ \ \  \ \ \ \ \ +\sum_{v=k}^{K}N_{\sigma_k}D_{KL}(N(0,{\sigma_k}^2)||N(0,{\sigma_k^v}^2))\},
\end{aligned}\label{obj_each}
\end{equation}
which is easily understandable to link all views of MoG noise parameters to a latent common one with parameter $\boldsymbol{\pi}$ and $\boldsymbol{\sigma}$. This KL-divergence term corresponds to an improper prior\footnote{Improper prior distribution~\cite{christensen2010improperprior} means that it does not integrate to 1.} imposed on local MoG parameters $\left \{ \mathbf{\Pi} ^{v} \right \}_{v=1}^V$ and $\left \{ \mathbf{\Sigma} ^{v} \right \}_{v=1}^V$. The MAP model (\ref{MAP}) is thus theoretically sound. More theoretical explanations on the model will be presented in Section 5.

Note that the physical meanings of the objective function (\ref{obj_each}) can be easily interpretted. The first term is the likelihood term, mainly aiming to fit input data $\mathbf{X}$, the second one encodes the prior knowledge on deterministic parameters $\boldsymbol{\theta}^{model}$, and the third one regularizes each stochastic variable $\boldsymbol{\theta}^{MoG}$ involved in the model by pulling it close to a latent shared distribution.

\section{EM algorithm for solving MSL-RMoG}

The EM algorithm is readily applied to solve (\ref{obj_fun}).
The algorithm contains three steps:
calculate the expectation of posterior of the latent variable $\mathbf{z}_{ij}^{v}$; optimize the
MoG noise parameters; optimize the model parameters.

\textbf{E Step:} the posterior responsibility of mixture component $k$ can be calculate  by
\vspace{-0pt}
\begin{equation}\small
\begin{aligned}
E(z_{ijk}^{v})=\gamma _{ijk}^{v}=\frac{\pi _{k}^{v}\mathcal{N}(e_{ij}^{v}\mid 0,{\sigma _{k}^v}^{2})}{\sum_{k=1}^{K}\pi _{k}^{v}\mathcal{N}(e_{ij}^{v}\mid ,0,{\sigma _{k}^v}^{2})}
\end{aligned}\label{E_step}
\end{equation}
We can then do the M-step by maximizing the corresponding upper bound w.r.t $\boldsymbol{\theta}^{model}$ and $\boldsymbol{\theta}^{MoG}$:
\vspace{-0pt}
\begin{equation}\small
\begin{aligned}
\mathcal{L'}(\boldsymbol{\theta})=
-E_{Z}\{\ln p(\mathbf{X},\mathbf{Z}|\boldsymbol{\theta})\}+\mathcal{R}(\boldsymbol{\theta}^{model})+\mathcal{R}(\boldsymbol{\theta}^{MoG}),
\end{aligned}\label{obj_fun1}
\end{equation}
where
\vspace{-0pt}
\begin{equation}\footnotesize
\begin{aligned}
E_{Z}\{\ln p(\mathbf{X},\mathbf{Z}|\boldsymbol{\theta})\}=
\ \    \sum_{v=1}^{V} \sum_{i,j\in \Omega^{v}}\sum_{k}^{K} \gamma _{ijk}^{v}\ln \left \{  \pi _{k}^{v}\mathcal{N}(e_{ij}^{v}\mid 0,{\sigma_k^v}^2) \right \}.
\end{aligned}
\end{equation}
\textbf{M Step for updating MoG parameters $\boldsymbol{\theta}^{MoG}$\footnote{The inference is listed at A1 in supplementary material.}:}
A way to address problem (\ref{obj_fun1}) is to take derivatives w.r.t. all the MoG
parameters and set them to zeros. The updating formulations of these parameters are written as follows:

\textbf{Update $\mathbf{\Pi }^{v}$ and $\mathbf{\Sigma }^{v}$:} Referring to \cite{dempster1977maximum}, the closed-form updating equations of $\mathbf{\Pi }^{v}$ and $\mathbf{\Sigma }^{v}$ are:
\vspace{-0pt}
\begin{equation}\small
\begin{aligned}
& N_{k}^v=\sum_{i,j}^{m,n}\gamma _{ijk}^{v}, \quad \pi _{k}^v=\frac{N_{k}^{v}+N_{\pi}\pi_k}{\sum_{k=1}^{K}(N_{k}^v+N_{\pi}\pi_k)},\\
& {\sigma _{k}^v}^{2}=\frac{\sum_{i,j}^{m,n}\gamma _{ijk}^{v}\{(e _{ij}^{v})^{2}+N_{\sigma_k}\sigma_k^2\}}{N_{k}^v+N_{\sigma_k}}.
\end{aligned}\label{M_step_MoG1}
\end{equation}

\textbf{Update $\mathbf{\Pi }$ and $\mathbf{\Sigma }$:}
The parameters $\mathbf{\Pi }$ and $\mathbf{\Sigma }$ also have a closed-form solution in M step:
\vspace{-0pt}
\begin{equation}\small
\begin{aligned}
\sigma_k^2=\frac{V}{\sum_{v=1}^{V}{\sigma_k^v}^{-2}};\ \
\pi_k=\frac{\sqrt[V]{\prod_{v=1}^V\pi_k^v}}{\sum_{k=1}^K\sqrt[V]{\prod_{v=1}^V\pi_k^v}}.
\end{aligned}\label{M_step_MoG2}
\end{equation}
Both parameters of the \mbox{common} MoG noise can be easily explained: $\sigma_k^2$ is the harmonic mean of $\{{\sigma_k^v}^{2}\}_{v=1}^V$ and $\pi_k$ is proportional to the geometric mean of $\{\pi_k^v\}_{v=1}^V$.

\textbf{M Step for model variables $\boldsymbol{\theta}^{model}$:}
The related terms in Eq. (\ref{obj_fun1}) can be equivalently reformulated as follows:
\begin{equation}\small
\min \sum_{v=1}^{V}\left \| \mathbf{W}^{v}\odot (\mathbf{X}^{v}-\mathbf{L}^{v}(\mathbf{R}^{v}+\mathbf{R})) \right \|_{F}^{2}+\mathcal{R}(\boldsymbol{\theta}^{model}),\label{M_step_model}
\end{equation}
where $\odot$ represents the Hadamard product and the element $w_{ij}^{v}$ of the indicator matrix $W^{v}$, with same size of $X^{v}$, is
\begin{equation}\small
w_{ij}^{v}=\left\{\begin{matrix}
\sqrt{\sum_{k=1}^{K}\frac{\gamma _{ijk}^{v}}{2 {\sigma _{k}^v}^{2}}}, &i,j\in \Omega^{v}  \\
0, &i,j\notin \Omega^{v}.
\end{matrix}\right.\label{W}
\end{equation}

There exist many off-the-shelf algorithms \cite{srebro2003weighted,buchanan2005damped,de2003framework}) to tackle Eq. (\ref{M_step_model}). We easily apply the ALS owing to its simplicity and effectiveness. The detailed steps of MSL-RMoG are then provided in Algorithm 1.

\begin{algorithm}[!tbp]\small
\caption{[MSL-RMoG]Multi-view subspace Learning with Regularized MoG}\label{alg1}
\begin{algorithmic}[1]
\renewcommand{\algorithmicrequire}{\textbf{Input:}}
\renewcommand{\algorithmicensure}{\textbf{End}}
\REQUIRE multi-view data: $\{\mathbf{X}^v\}_{v=1}^V$; subspace rank : $r$;\\ the number of Gaussians : $K$;
model parameters: $\lambda$, $\lambda _{R}$, $\lambda _{L}$
MoG parameters: $\{N_{\sigma_k}\}_{k=1}^K$ and $N_{\pi}$
\renewcommand{\algorithmicrequire}{\textbf{Initialization:}}
\renewcommand{\algorithmicensure}{\textbf{End}}
\REQUIRE $\boldsymbol{\theta}^{model}$ and $\boldsymbol{\theta}^{MoG}$
\WHILE {not converged}
\FOR{$v=1,2,...,V$}
\STATE
\textbf{E-step}: Evaluate $\gamma_{ijk}^{v}$ by Eq. (\ref{E_step})
\STATE
 \textbf{M-step for MoG parameters $\boldsymbol{\theta}^{MoG}$}:
 \\compute $\{\mathbf{\Pi}^{v}\}_{v=1}^V, \{\mathbf{\Sigma}^{v}\}_{v=1}^V, \{N^{v}\}_{v=1}^V,$ by Eq. (\ref{M_step_MoG1})
 \STATE
 \textbf{M-step for model varables $\boldsymbol{\theta}^{model}$}:
 \\compute $\mathbf{W}^{v}$by equation (\ref{W})
 \\compute $\{\mathbf{L}^{v}\}_{v=1}^V, \{\mathbf{R}^{v}\}_{v=1}^V$ by solving Eq. (\ref{M_step_model})\ \ \ \ \ \ \ \ \ \ \ \ \ \ \ \ \ \ through ALS
\ENDFOR
\STATE
compute $\mathbf{\Pi}, \mathbf{\Sigma}$ by Eq. (\ref{M_step_MoG2})\\
 compute $\mathbf{R}$
 by solving Eq.  (\ref{M_step_model}) through ALS

\ENDWHILE
\renewcommand{\algorithmicrequire}{\textbf{Output:}}
\renewcommand{\algorithmicensure}{\textbf{End}}
\REQUIRE $\{\mathbf{\Pi}^{v}\}_{v=1}^V, \{\mathbf{\Sigma}^{v}\}_{v=1}^V, \{\mathbf{L}^{v}\}_{v=1}^V, \{\mathbf{R}^{v}\}_{v=1}^V, \mathbf{R}$
\end{algorithmic}
\end{algorithm}

The memory consumption and complexity of MSL-RMoG are $O(mnVK+mrV+rnV)$ and $O(I_2(mnVK+I_{1}mrnV))$, where $I_{1}$ is iteration number of inner loop in ALS, and $I_{2}$ is iteration number for outside loop in MSL-RMoG. This complexity is comparable to or even less than those of the current MSL methods \cite{lock2013joint,guo2013convex,xu2015multi,white2012convex}. In our experiments, the setting of $I_{1}$ is not sensitive to the algorithm performance. We just empirically specify it as a small number.

\section{KL divergence regularization}
\subsection{Theoretical explanation}
\textbf{Relationship to conjugate prior:}
For MoG parameters $\left\{\mathbf{\Pi} ^{v} \right \}_{v=1}^V$, $\left \{ \mathbf{\Sigma} ^{v} \right \}_{v=1}^V$ defined  in each view, the KL regularization term in Eq. (\ref{obj_each}) can be explained
from the perspective of conjugate prior. ~\cite{yong2017robust} shows that the relationship between
KL divergence and conjugate prior under fully  exponential family distribution. In this paper, we
can show that this conclusion is also correct for all exponential family distributions. The theorem can
be summarized as:

\textbf{Theorem 1}
\emph{If a distribution $p(\mathbf{x}|\boldsymbol{\theta})$ belongs to the  exponential family with the form:
$p(\mathbf{x}|\boldsymbol{\theta})=h(\mathbf{x})\eta(\boldsymbol{\theta})\text{exp}
({\boldsymbol{\theta}}^T\boldsymbol{\phi}(\mathbf{x}))$ with natural parameter $\boldsymbol{\theta}$,
and its conjugate prior follows: $\small{p(\boldsymbol{\theta}|\boldsymbol{\mathcal{X}},\gamma)
=f(\boldsymbol{\mathcal{X}},\gamma)\eta(\boldsymbol{\theta})^\gamma\text{exp}(\gamma
{\boldsymbol{\theta}}^T\boldsymbol{\mathcal{X}})},$
then we have:
\vspace{-0pt}
$$\ln p(\boldsymbol{\theta}|\boldsymbol{\mathcal{X}},\gamma)=
-\gamma D_{KL}(p(\mathbf{x}|\boldsymbol{\theta}^{*})||p(\mathbf{x}|\boldsymbol{\theta}))+C,$$
where $\boldsymbol{\theta}^{*}=arg\max_{\boldsymbol{\theta}}{p(\boldsymbol{\theta}|\boldsymbol{\mathcal{X}},\gamma)}$
and $C$ is a constant independent of $\boldsymbol{\theta}$}.

Specifically, by giving $\mathbf{\Pi}^{v}$ a
Dirichlet distribution prior and ${{\sigma}_k^{v}}^2$ Inverse-Gamma  distribution prior as:
\vspace{-0pt}
$$p(\mathbf{\Pi}^{v})=\text{Dir}(\mathbf{\Pi}^{v}|\{N_{\pi}\pi_k+1\}_{k=1}^K),$$
$$p({{\sigma}_k^{v}}^2)=\text{Inv-Gamma}\left ({{\sigma}_k^{v}}^2|\frac{N_{\sigma_k}}{2}-1,\frac{N_{\sigma_k}{{\sigma}_k}^2}{2}\right ),$$
we can deduce the KL-divergence regularization terms in Eq.(\ref{obj_each}) for $\{\mathbf{\Pi}^{v}\}_{v=1}^V$ and $\{\mathbf{\Sigma}^{v}\}_{v=1}^V$ under MAP framework.

\textbf{Theorem 2}
\emph{If $p(\mathbf{x}|\boldsymbol{\theta}_1)$ and $p(\mathbf{x}|\boldsymbol{\theta}_2)$ are the same type of exponential family distribution $p(\mathbf{x}|\boldsymbol{\theta})=h(\mathbf{x})\eta(\boldsymbol{\theta})\text{exp}
({\boldsymbol{\theta}}^T\boldsymbol{\phi}(\mathbf{x}))$, then $$D_{KL}(p(\mathbf{x}|\boldsymbol{\theta}_1)||p(\mathbf{x}|\boldsymbol{\theta}_2))=B_F(\boldsymbol{\theta}_2||\boldsymbol{\theta}_1),$$
where $B_F(\cdot||\cdot)$ is the Bregman divergence with convex function $F(\boldsymbol{\theta})=-\log(\eta(\boldsymbol{\theta}))$, which can be defined as $B_F(\boldsymbol{\theta}_1||\boldsymbol{\theta}_2)=F(\boldsymbol{\theta}_1)-F(\boldsymbol{\theta}_2)-
\nabla F(\boldsymbol{\theta}_2)^T(\boldsymbol{\theta}_1-\boldsymbol{\theta}_2)$.
}

 ~\cite{nielsen2009statistical} has proven this conclusion. Instead of calculating function integral, this Theorem can give a fast solution to calculate the KL divergence between two same type exponential family distributions.

\textbf{KL divergence average:} For the parameters $\{\mathbf{\Pi},\mathbf{\Sigma}\}$ defined in the shared latent noise distribution, it corresponds to a KL divergence average problem. Specifically, we can prove the following theorem:

\begin{table*}\small
\vspace{-0mm}
\caption{Performance comparison of different methods on Multi-PIE data in terms of PSNR}
\centering
\begin{tabular}{c |c c c| c c c |c c c}
\shline
{Noise type}&\multicolumn{3}{c}{Gaussian}&\multicolumn{3}{c}{Sparse}&\multicolumn{3}{c}{Mixture}\\
\hline
{View}&$-15^{\circ}$&$0^{\circ}$ & $15^{\circ}$&$-15^{\circ}$&$0^{\circ}$ & $15^{\circ}$&$-15^{\circ}$&$0^{\circ}$ & $15^{\circ}$\\
\hline
{Noise image}&14.01&13.95&13.96 &11.82&11.67&11.73&3.62&3.75&3.63\\
{JIVE}&22.69&22.72&22.75&21.19 &21.37&21.49&14.15& 14.15 & 14.24\\
{CSRL}&23.14&\emph{23.38}&\emph{23.03}&21.79 &22.05&21.74&14.37&14.33 &14.41\\
{MISL}&\emph{23.17}&23.24&22.98 &22.03&22.26&21.88&19.74&19.56&19.15\\
{MSL}&21.30&21.59&21.31&\textbf{25.90}&\textbf{26.89}&\emph{25.83}&\emph{25.04}&\emph{25.95}&\emph{25.48}\\
{MSL-RMoG}&\textbf{23.20}&\textbf{23.65}&\textbf{23.37}
&\emph{25.77}&\emph{26.69}&\textbf{25.98}&\textbf{25.40}&\textbf{26.56}&\textbf{25.83}\\
\hline
\end{tabular}\vspace{-0mm}
\label{psnr}
\end{table*}

\textbf{Theorem 3}
\emph{If distributions $p(\mathbf{x}|\boldsymbol{\theta})$ and $p(\mathbf{x}|\boldsymbol{\theta}^v)(v=1,2,...,V)$,  belong to the same kind of full exponential family distribution,
which means they all have the form:
$p(\mathbf{x}|\boldsymbol{\theta})=\eta(\boldsymbol{\theta})\text{exp}
({\boldsymbol{\theta}}^T\boldsymbol{\phi}(\mathbf{x})),$
then  the solution of the problem:
$\min_{\boldsymbol{\theta}} \sum_{v=1}^V D_{KL}(p(\mathbf{x}|\boldsymbol{\theta})||p(\mathbf{x}|\boldsymbol{\theta}^{v}))
$
is $
\boldsymbol{\theta}=\frac{1}{V} \sum_{v=1}^V \boldsymbol{\theta}^{v}.
$
}

For instance, the natural parameter of Gaussian distribution  $N(x|0,{\sigma_k^v}^2)$ is ${\sigma_k^v}^{-2}$, so their KL divergence average is ${\sigma_k}^{-2}=\frac{1}{V}{\sigma_k^v}^{-2}$, which can lead to the result in Eq.(\ref{M_step_MoG2}).

\subsection{ Joint regularization for mixture distribution}
Generally speaking, the mixture of full exponential family distributions does not belong to the fully exponential family, so for this mixture distribution we can use a independent KL divergence for each parameter of its distribution like Eq. (\ref{obj_each}).
 However, there are so many  hyperparameters to be set with this approach.
Actually,
we can further prove that the joint distribution $p(\mathbf{x},\mathbf{z}|\boldsymbol{\theta})$  of observed variables $\mathbf{x}$ and latent variable $\mathbf{z}$ is exactly a fully exponential family distribution:

 \textbf{Theorem 4}
\emph{If the distributions $p(\mathbf{x}| z_k=1),(k=1,2,...K)$ all belong to the full exponential family with natural parameter $\boldsymbol{\theta}_k$
and $p(\mathbf{z})=\text{Multi}(\mathbf{z}|\boldsymbol{\pi})$
($z_k\in \{0,1\}$,$0\leq \pi_k\leq1 $ and $ \sum_{k=1}^K\pi_k=1$),
then, $p(\mathbf{x},\mathbf{z})$ belongs  to the  exponential family.
}

Moreover, we can also prove:

\textbf{Theorem 5}
\emph{ For any two distributions $p(\mathbf{x},\mathbf{z}|\boldsymbol{\theta}_1)$ and
$p(\mathbf{x},\mathbf{z}|\boldsymbol{\theta}_2)$ with their marginal distributions $p(\mathbf{x}|\boldsymbol{\theta}_1))$and $p(\mathbf{x}|\boldsymbol{\theta}_2))$, the following inequality holds
$$D_{KL}(p(\mathbf{x}|\boldsymbol{\theta}_1)
||p(\mathbf{x}|\boldsymbol{\theta}_2))\leq D_{KL}(p(\mathbf{x},\mathbf{z}|\boldsymbol{\theta}_1)
||p(\mathbf{x},\mathbf{z}|\boldsymbol{\theta}_2)).
$$
}
\vspace{-0mm}

From the theorem, it is easy to see that $D_{KL}(p(\mathbf{x},\mathbf{z}|\{\boldsymbol{\theta}_k\}_{k=1}^K,\boldsymbol{\pi})
||p(\mathbf{x},\mathbf{z}|\{\boldsymbol{\theta}_k^v\}_{k=1}^K,\boldsymbol{\pi}^v))$ constitutes an upper bound of the KL divergence between original two mixture distributions, and thus can be rationally used as a regularization term for the original mixture distribution.
For example, the joint KL divergence regularization for MoG is
\begin{equation}\small
\mathcal{R}(\boldsymbol{\theta}^{MoG})=N D_{KL}(p(x,\mathbf{z}|\mathbf{\Pi}^{v},\mathbf{\Sigma}^{v})||p(x,\mathbf{z}|\mathbf{\Pi},\mathbf{\Sigma})),
\label{Joint_KL}
\end{equation}
where
$\footnotesize{p(x,\mathbf{z}|\mathbf{\Pi},\mathbf{\Sigma})=\prod_{k=1}^K{{\pi_k}^{z_k}\mathcal{N}(x|0,{\sigma_k}^2)^{z_k}}}$. This regularization term also leads to a simple solution, by setting $N_{\pi}=N$ and $N_{\sigma_k}=N\pi_k$ in Eq. (\ref{M_step_MoG1}) and let
\vspace{-0pt}
$$
\pi_k=\frac{\sqrt[V]{\prod_{v=1}^V\pi_k^ve^{-D_{KL}(N(0,{\sigma_k}^2)||N(0,{\sigma_k^v}^2))}}}
{\sum_{k=1}^K\sqrt[V]{\prod_{v=1}^V\pi_k^ve^{-D_{KL}(N(0,{\sigma_k}^2)||N(0,{\sigma_k^v}^2))}}}.
$$
The solution for other variables and parameters is not changed. The joint KL divergence regularization only have one compromising parameter $N$ to be set, which can be generally easy to set.
The proofs of all theorems are listed in supplementary material.

\begin{figure*}
  \centering
  \vspace{-0mm}
  \includegraphics[width=140mm]{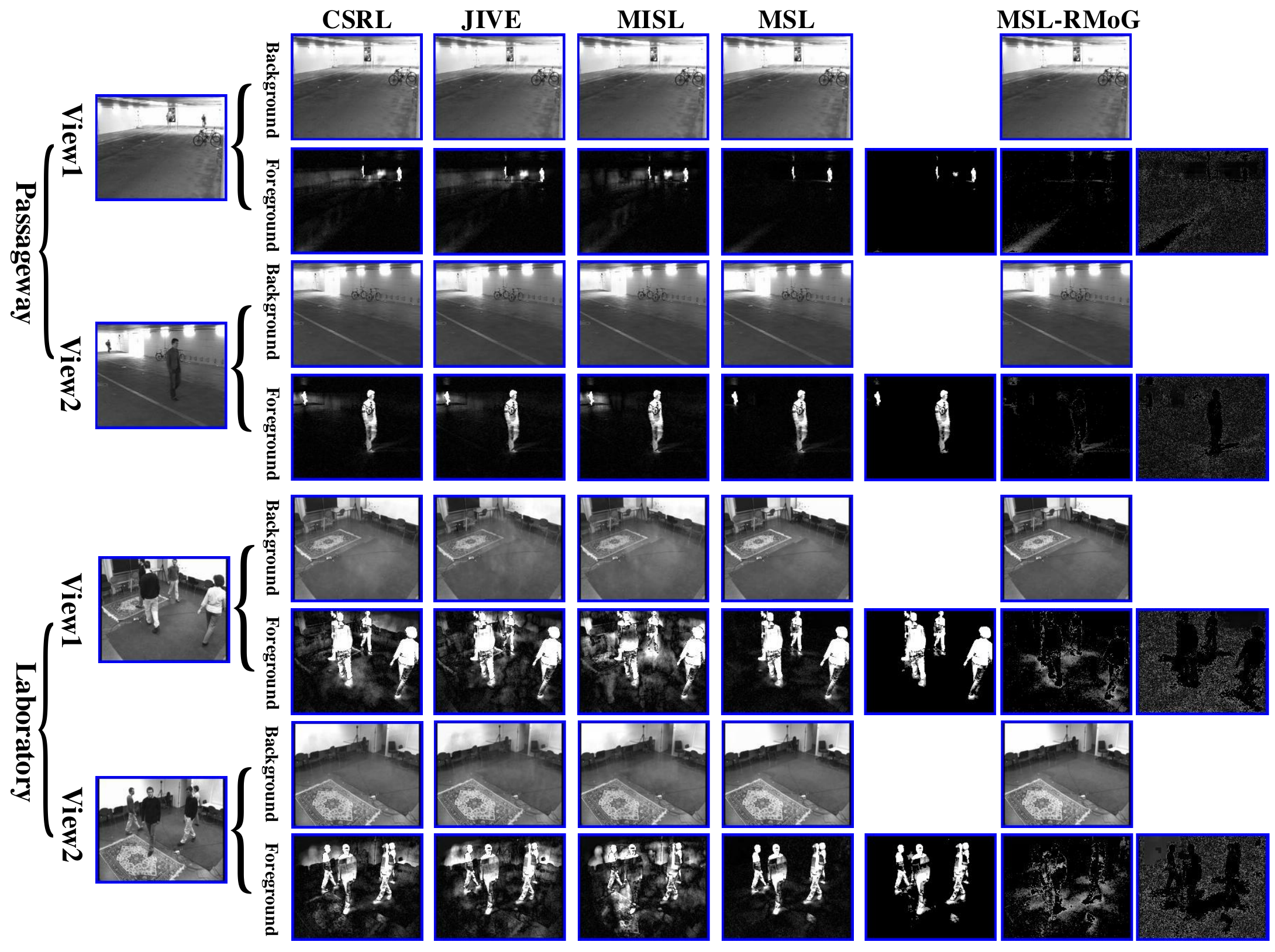}
  \vspace{-0mm}
  \caption{Results of background subtraction in multi-view passageway and laboratory videos. The residuals of MSL-RMoG can be divided into three Gaussians components with different scales, as shown in $6^{th}$ to $8^{th}$ columns (after scale processing) .}
  \vspace{-0mm}
  \label{fig:mog_fore}
\end{figure*}

\section{Experimental results}

\begin{figure}[!]
  \centering
  \vspace{-0mm}
  \includegraphics[width=87mm]{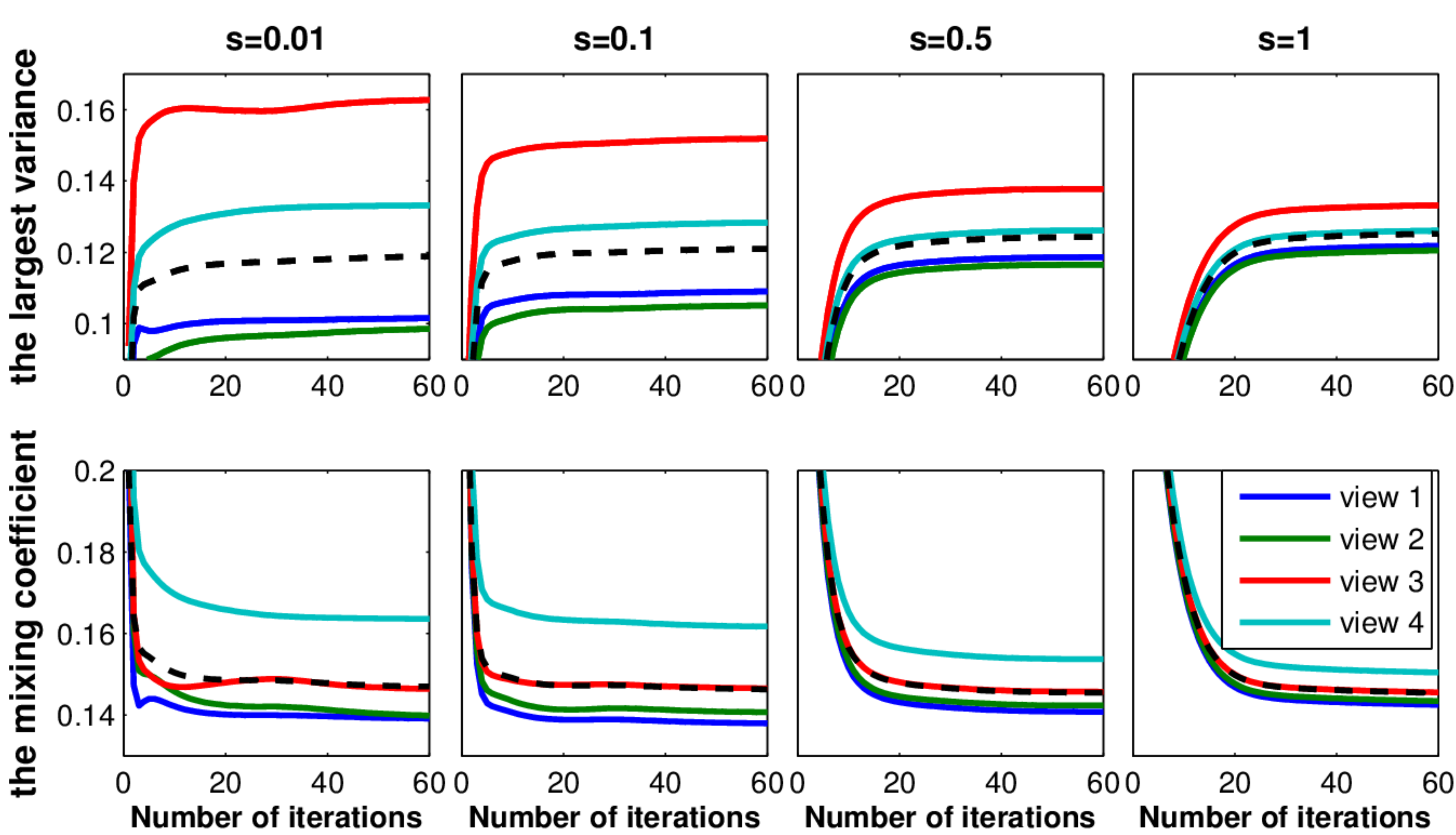}
  \vspace{-0mm}
  \caption{The curves of $\sigma_1^v$ and
 $\pi_1^v$ with different KL strength parameters
}
  \vspace{-0mm}
  \label{LLH_1}
\end{figure}

To qualitatively and quantitatively evaluate the performance of our proposed multi-view subspace Learning with the complex noise method, we conduct three types of experiments containing face image reconstruction, multi-view and RGB background subtractions. We compare our method with JIVE \cite{lock2013joint}, CSRL \cite{guo2013convex}, MISL\cite{xu2015multi} and MSL \cite{white2012convex}, which represent the state-of-the-art MSL development. Most parameters of the compared methods is set to be the default value and the rank is set the same for all these methods. For MSL-RMoG, the number of MoG components is set as 3 in all cases, except 2 in face experiments with Gaussian noise. The model parameters $\lambda_L$ ,$\lambda$ and $\lambda_R$ and are set as
$0.001$,$0.001$ and $1$, respectively. Besides, we use joint regularization in Eq. (\ref{Joint_KL})
with strength parameter $N=0.2mnV/K$ throughout all our experiments.

\subsection{Background Subtraction on Multi-view data}
In this experiment, our method is applied to the problem of background subtraction. Two multi-camera pedestrian videos \cite{Berclaz11,Fleuret08a}, shot by 4 cameras located at different angles, are employed: Passageway and Laboratory. All the frames in the video are 288$\times$360. Without loss of generality, we resize the original frames with 144$\times$180. 200 frames of Passageway and Laboratory sequences beginning at the first frame and ending at the 1000$^{th}$ frame (take the first one of each 5 frames) are extracted to compose the learning data.
The JIVE, CSRL, MISL, MSL and our MSL-RMoG method are implemented in these videos, and the rank is set as 2 for all methods. Fig. \ref{fig:mog_fore}
shows the results obtained by all competing methods on some typical frames on the multi-view video data.

From the figure, it is seen that the background image achieved by our proposed method is clearer in details. Compared with most other competing methods, the MSL-RMoG method is able to extract the foreground objects in a more accurate manner. As shown in Fig. \ref{fig:mog_fore}, MSL-RMoG decomposes the foreground into three components with different extents, each having its own physical meaning: (1) moving objects in the foreground; (2) shadows alongside the foreground objects due to lighting changes and people walking; (3) background variation caused mostly by the camera noise. As most existing methods merge the object, its shadow and background noise, the foreground extracted by them is relatively more coarse. Besides, in order to illustrate how the KL divergence regularization works, we set the strength parameter $N=smnV/K$ with different $s$ in Laboratory video experiments and draw the curves of largest variance $\sigma_1^v$ and its
mixing coefficient $\pi_1^v$ in Fig. \ref{LLH_1}. we can
find  the KL divergence regularization  makes  the distributions of noise in different views have a certain similarity.

\begin{figure*}
  \centering
  \vspace{-0mm}
  \includegraphics[width=140mm]{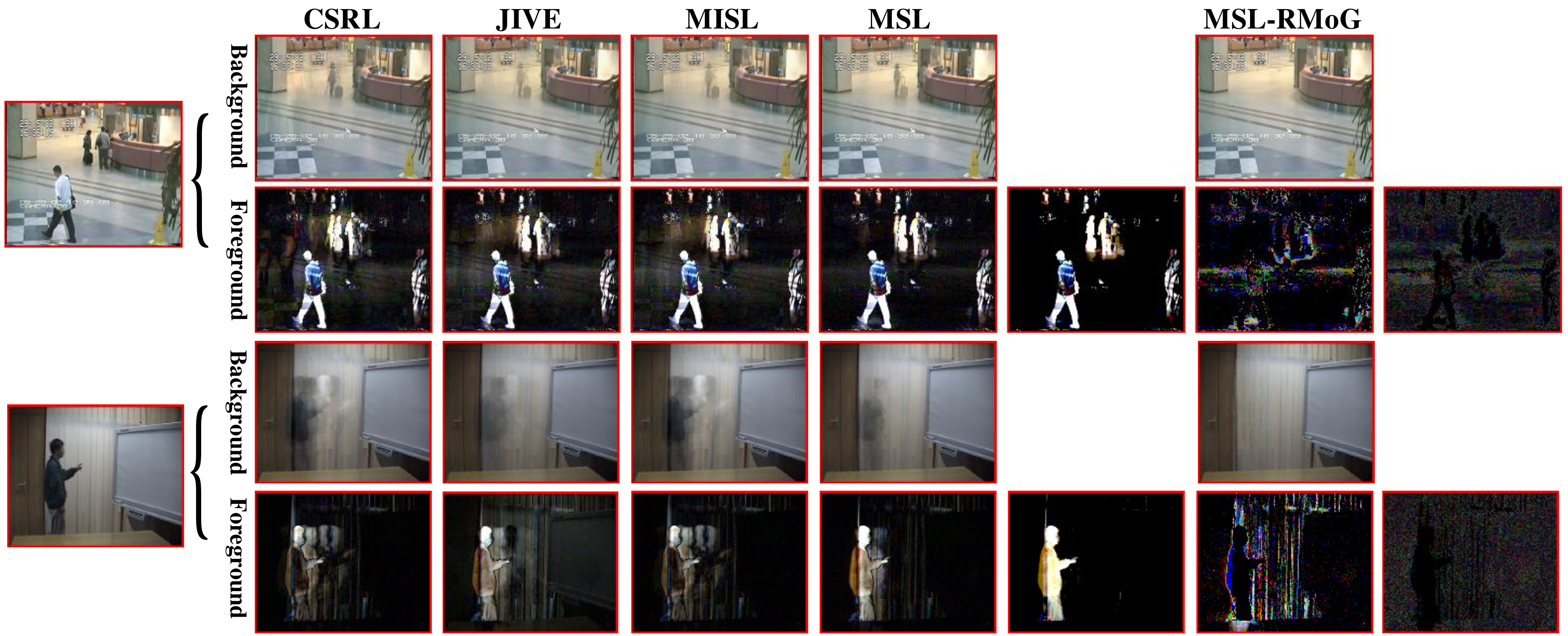}
  \vspace{-0mm}
  \caption{Results of background subtraction in two RGB video data: airport and curtain.}
  \vspace{-0mm}
  \label{fig:rgb_fore}
\end{figure*}

\subsection{Face Images Recovery Experiments}

\begin{figure}[!]
  \centering
  \vspace{-0mm}
  \includegraphics[width=76mm]{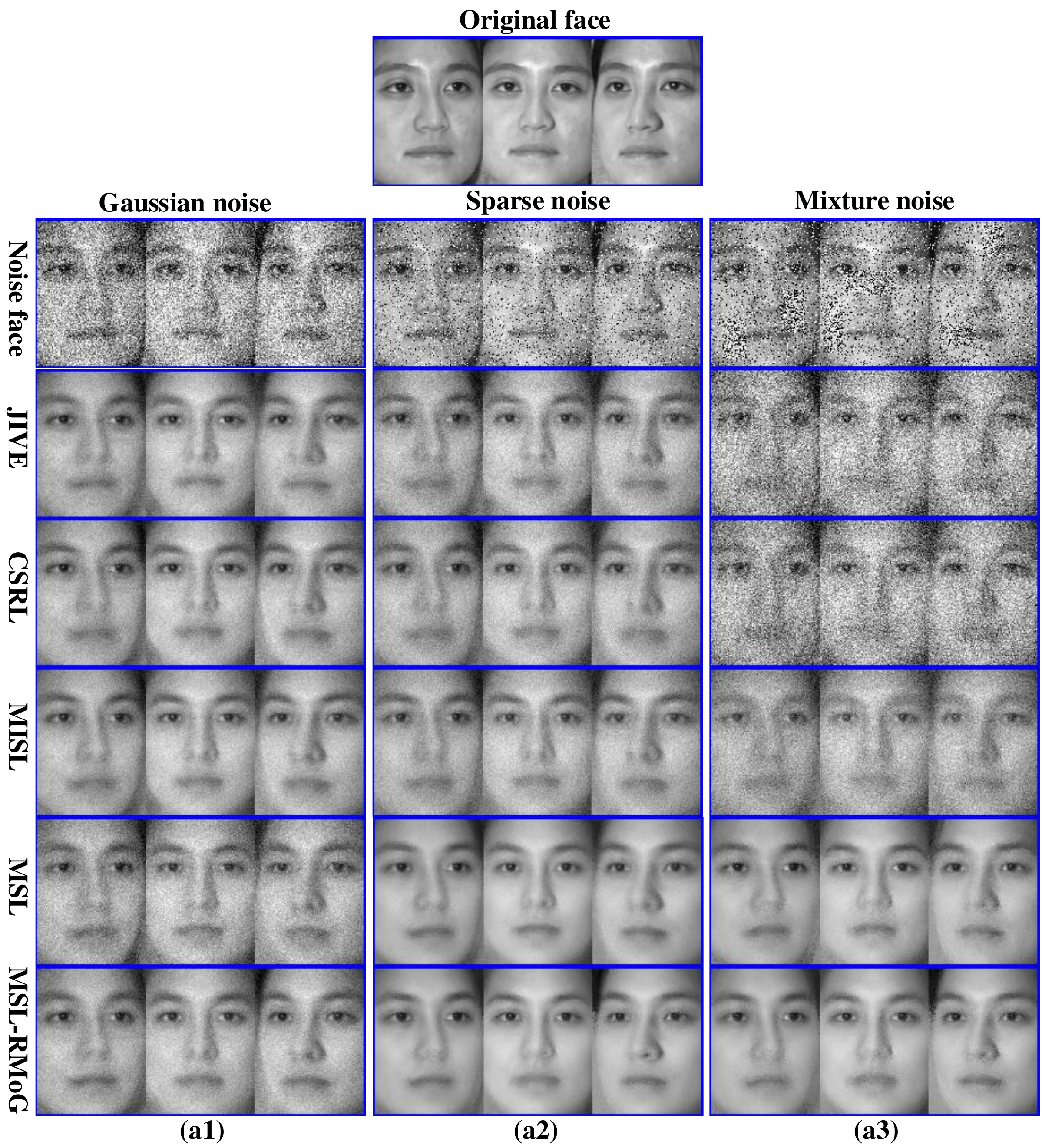}
  \vspace{-0mm}
  \caption{Reconstructed face images of a typical subject in the CMU Multi-PIE face dataset by different competing methods.
}
  \vspace{-0mm}
  \label{fig:mog_face}
\end{figure}

This experiment aims to test the effectiveness of the proposed MSL-RMoG methods in face images reconstruction. The CMU Multi-PIE face dataset \cite{gross2010multi} including 337 subjects with size 128$\times$96 is used, in which images are multiple poses and expressions. 200 subjects are randomly selected and each subject contains 3 views ( $-15^{\circ}$, $0^{\circ}$ ,$15^{\circ}$ ). In this experiment, we add different types of noise or outliers to the original image: (1) \emph{Gaussian noise} $\mathcal{N}$(0,0.15) (shown in the $1^{st}$ row of Fig. \ref{fig:mog_face} (a1)); (2) \emph{sparse noise}: random 20$\%$ $\mathcal{U}(-1,1)$ noise (shown in the $1^{st}$ row of Fig. \ref{fig:mog_face}(a2)); (3) \emph{mixture noise}: Gaussian noise $\mathcal{N}$(0,0.02)+block occlusion (salt\&pepper noise inside)+20$\%$ sparse noise (shown in the $1^{st}$ row of Fig. \ref{fig:mog_face}(a3)). The comparison methods include JIVE, CSRL, MISL, and MSL, and the rank is set as $20$ in this experiments. Different views of reconstructed images obtained by different methods are shown in Fig. \ref{fig:mog_face}. The PSNR values of images obtained by different methods are listed in Table \ref{psnr}.

From the figure and table, it is easy to observe that our proposed MSL-RMoG method is capable of finely recovering the clean faces in various noise cases,
especially in the case of relatively more complicated mixture noises. MSL  performs well on sparse noise, but not well on
Gaussian noise. JIVE and CSRL can not work well on the sparse noise and mixture noise since they implicitly assume the noise as a i.i.d. Gaussian. MISL performs slightly better than JIVE and CSRL as a result of the utilization of Cauchy loss, which is more robust than $L_2$ loss.

\begin{table}\footnotesize
\caption{Performance comparison of all competing methods on all video sequences in Li dataset in terms of F-measure.
}
\centering
\begin{tabular}{c c c c c c}
\hline
\multirow{2}{*}{Video}&\multicolumn{5}{c}{Methods}\\
\cline{2-6}&JIVE&CSRL&MISL&MSL& MSL-RMoG\\
\hline
air.&0.6450    &0.6476    &0.6536   & 0.6722  &  \textbf{0.6770}\\
boo.& 0.6553  &  0.6622    &0.6619  &  \textbf{0.6902}  &  0.6778\\
sho.    & 0.7145  &  0.7186   & 0.7205  &  \textbf{0.7274}  &  0.7264\\
lob.  &  0.5426  &  0.5442   & 0.5444   & 0.6771  &  \textbf{0.7781}\\
esc.& 0.5911  &  0.5947   & 0.5945   & \textbf{0.6142} &   0.5994\\
cur.& 0.5083   & 0.5393   & 0.5403   & 0.6198  &  \textbf{0.7189}\\
cam.&0.4186   & 0.4190    &0.4199   & \textbf{0.4609}  &  0.4405\\
wat.&  0.6028 &   0.6026  &  0.6029  &  0.8647   &  \textbf{0.8709}\\
fou.&  0.5857  &  0.5895   & 0.6341   & 0.7148  &  \textbf{0.7181}\\
\hline
Average  &  0.5849 &   0.5908   & 0.5969  &  0.6713  &\textbf{0.6897}\\
\hline
\end{tabular}
\label{f-measure}
\vspace{-0mm}
\end{table}

\subsection{Background Subtraction on RGB data}
MSL-RMoG is further applied to the background subtraction on RGB data. In this experiment, we regard the three channels (red, green and blue) of a video as its three different views.
Actually different channels do have different residual distributions since the sensitive spectral band of R,G and B sensors is distinct. However, it is
the same objects that all sensors observed, therefore, residual distributions also have a certain degree of similarity. The noises in this data are thus with a more complicated non-i.i.d. structures than those assumed by the current MSL methods.

Fig. \ref{fig:rgb_fore} shows the result of some typical frames on the Li dataset with all competing methods including JIVE, CSRL, MISL, MSL and MSL-RMoG. We can easily observe that MSL-RMoG achieves clearer background, and meanwhile, the extracted foreground objects by our method is also of a better visualization effect.
Table \ref{f-measure} shows the F-measure of all methods in Li data set, which quantitatively shows the better performance of the proposed method.

\section{Conclusion}
\vspace{-0pt}
Current multi-view learning (MSL) methods mainly emphasize deterministic shared knowledge of data, but ignore the complexity, non-consistency and similarity of noise. This, however, is deviated from most real cases with more complicated noises in-between views and alleviates their robustness in practice. This paper has proposed a new MSL method, which firstly investigates this MSL noise issue and formulates the model capable of both adapting intra-view noise complexity (by parametric MoG) and delivering inter-view noise correlation (by KL-divergence regularization). Its novelty reflects in both its investigated issue and designed methodology (regularized noise modeling). Further, we also give a detailed and reasonable theoretical explanation for this term.
Experiments show that the new method is potentially useful to compensate previous MSL research to further enhance performance in complex noise scenarios.

%
\IEEEpeerreviewmaketitle

\ifCLASSOPTIONcaptionsoff
  \newpage
\fi




\bibliographystyle{IEEEtran}
\end{document}